\documentclass[letterpaper, 10 pt, conference]{ieeeconf}  

\IEEEoverridecommandlockouts                              

\usepackage{lipsum}
\usepackage{color}
\usepackage[colorinlistoftodos]{todonotes}

\usepackage{algorithm}
\usepackage{algpseudocode}
\usepackage{pifont}

\usepackage{amsmath}
\usepackage{amssymb}
\usepackage{amsfonts} 
\usepackage{mathtools}
\usepackage{bm}
\usepackage{mathrsfs}
\usepackage{eucal} 

\title{\LARGE \bf
Safe Reinforcement Learning on Autonomous Vehicles
}

\author{David Isele, Alireza Nakhaei, and  Kikuo Fujimura\\
Honda Research Institute USA \\
{\tt\small \{disele, anakhaei, kfujimura\}@honda-ri.com}%
}

\begin{document}

\maketitle
\thispagestyle{empty}
\pagestyle{empty}

\begin{abstract}  

There have been numerous advances in reinforcement learning, but the typically unconstrained exploration of the learning process prevents the adoption of these methods in many safety critical applications. Recent work in safe reinforcement learning uses idealized models to achieve their guarantees, but these models do not easily accommodate the stochasticity or high-dimensionality of real world systems. We investigate how prediction provides a general and intuitive framework to constraint exploration, and show how it can be used to safely learn intersection handling behaviors on an autonomous vehicle.

\end{abstract}


\section{INTRODUCTION}

With the increasing complexity of robotic systems, and the continued advances in machine learning, it can be tempting to apply reinforcement learning (RL) to challenging control problems. However the trial and error searches typical to RL methods are not appropriate to physical systems which act in the real world where failure cases result in real consequences. 


To mitigate the safety concerns associated with training an RL agent, there have been various efforts at designing learning processes with safe exploration. As noted by Garcia and Fernandez \cite{garcia2015comprehensive}, these approaches can be broadly classified into approaches that modify the objective function and approaches that constrain the search space.

Modifying the objective function mostly focuses on catastrophic rare events which do not necessarily have a large impact on the expected return over many trials. Proposed methods take into account the variance of return \cite{howard1972risk}, the worst-outcome \cite{heger1994consideration,howard1972risk,lipton2016combating}, and the probability of visiting error states \cite{geibel2005risk}. Modified objective functions may be useful on robotic systems where a small number of failures are acceptable. However on safety critical systems, often a single failure is prohibited and learning must be confined to \emph{always} satisfy the safety constraints. 

For this reason methods that constrain the search space are often preferable. Because these approaches can completely forbid undesirable states, they are usually accompanied by formal guarantees, however satisfying the necessary conditions on physical systems can be quite difficult in practice.  
For example, strategies have assumed a known safe policy which can take over and return to safe operating conditions\cite{hans2008safe}, a learning model that is restricted to tabular RL methods \cite{wen2015correct,wen2016probably}, and states that can be deterministically perceived and mapped to logical expressions \cite{alshiekh2018safe,shalev2016safe}. 

\begin{figure}[t]
\centering
\includegraphics[height=1.75in]{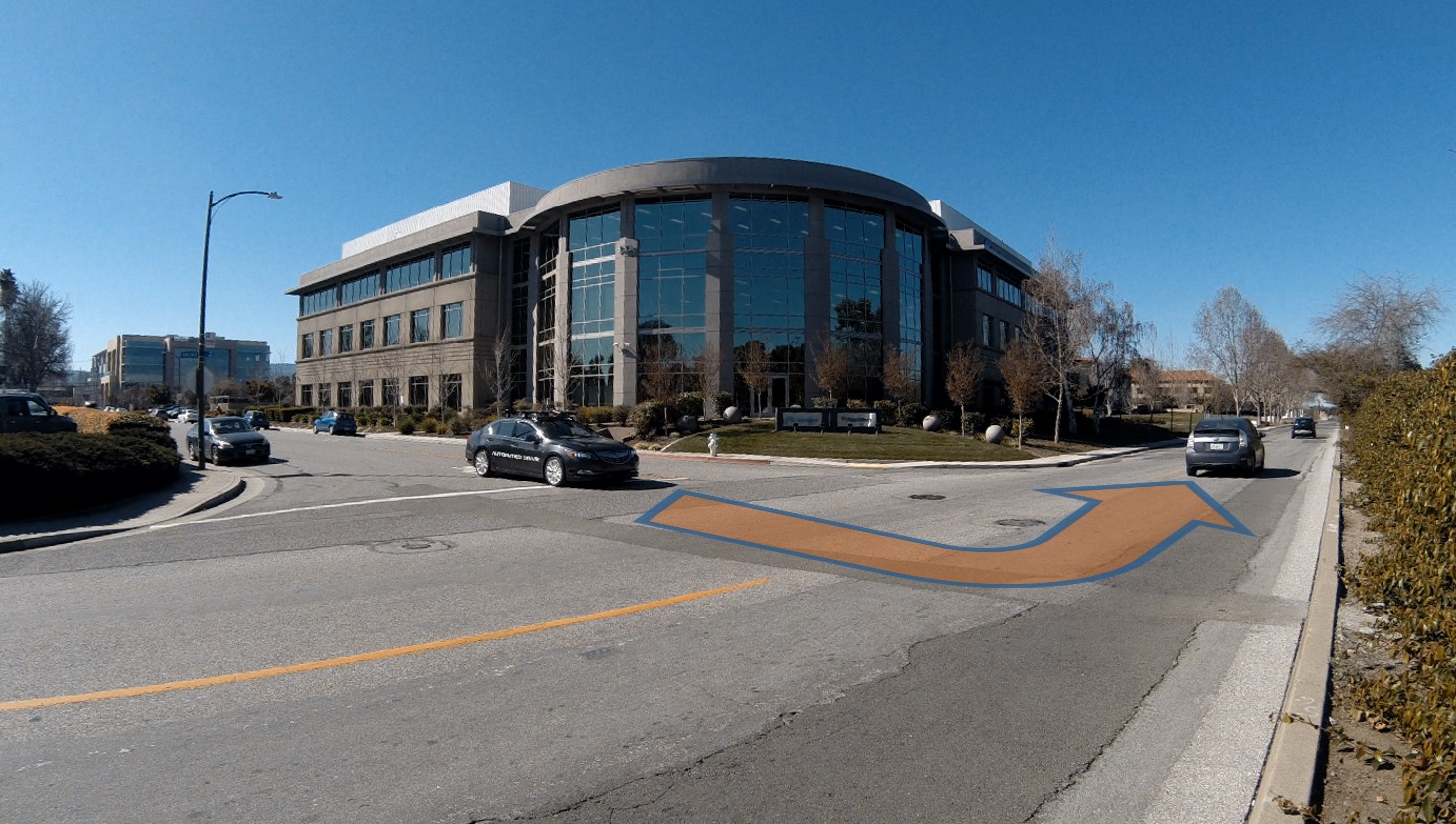}
\caption{An autonomous vehicle navigating an intersection. Prediction is used to shield the vehicle from making dangerous decisions, while allowing it to learn policies that are both efficient and not disruptive to other vehicles.}\label{fig:car}
\end{figure}

While there are some approaches that have been implemented on physical robots such as the work of Gillula et al. which uses reachability to enforce strict safety guarantees \cite{gillula2013reducing}, these approaches tend to be computationally expensive, preventing their application to high dimensional problems such as domains with multiple agents.

We investigate how prediction can be used to achieve a system that scales better to higher dimensions and is more suited to noisy measurements. Using prediction methods we show that we can safely constrain learning to optimize intersection behaviors on an autonomous vehicle where it must consider the behaviors of multiple other agents.  
While we believe prediction is a very general framework that lends itself to implementations on a variety of stochastic physical systems, we note that its safety constraints are weaker than other approaches in the literature:
we assume other agents (traffic vehicles) follow a distribution and are not adversarial. 

\begin{figure}
\centering
\vspace{5pt}
\includegraphics[height=3.75in]
{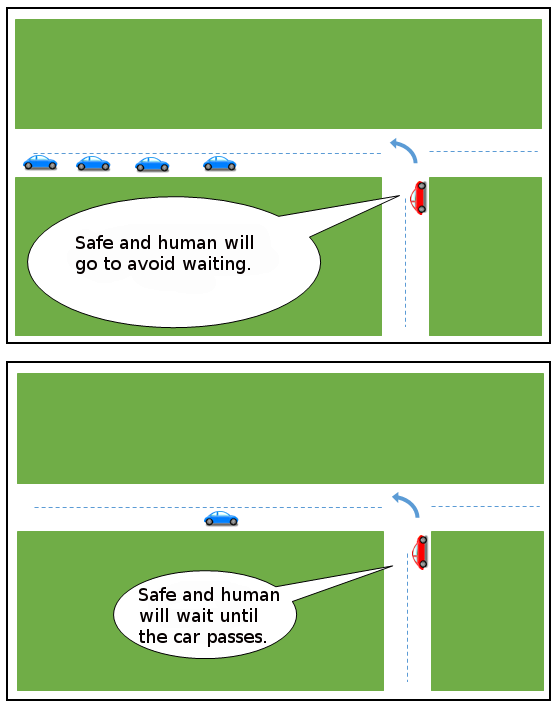}
\caption{In dense traffic, a human driver (red vehicle) might take the opening to minimize the wait that would result from the approaching heavy traffic. However, in sparse traffic, it would be more preferable to accept a small delay and let the car pass.}\label{fig:case_study}
\end{figure}

The specific application we investigate is making a turn at an unsigned intersection. This problem was recently explored as a non-safety constrained RL domain \cite{isele2018icra} where it was noted that the learned policy, which optimized efficiency, might be disruptive to traffic vehicles in practice. The primary concerns of these maneuvers are safety and efficiency, but balancing the two is a dynamic task. In dense traffic we may wish to seize an opportunity that leaves only several meters of a safety margin, as we might have to wait an unacceptable amount of time for the next opportunity. However in sparse traffic, adding an increased margin will only add a negligible delay and will likely be preferable to the passengers and other traffic vehicles. Figure \ref{fig:case_study} demonstrates the scenarios in dense and sparse traffic. 

We demonstrate the use of prediction as a safety constraint by learning a policy that minimizes disruption to traffic (as measured by traffic braking) while avoiding collisions. Additionally, we learn a policy that maximizes distance to other vehicles, while still getting through the intersection in a fixed time window. We show that these two optimizations produce different behaviors and that both can be learned using RL with 0 collisions.

\section{PROBLEM STATEMENT}

In this document we use the subscript/superscript notation $variable_{time}^{agent,action}$.
We define a safe set of policies $\bm{\Pi}^{i}$ as the set of policies $\pi$ that 
generates a trajectory $\tau$ that with probability less than $\delta$ has agent $i$ entering a danger state at any step in its execution\footnote{Interesting corner cases were proposed for many existing definitions of safety by Moldovan and Abbeel \cite{moldovan2012safe}. Their proposed definition of safety in terms of ergodicity does not easily extend to a multi-agent setting.}.

To find a safe policy in a multi-agent setting, we formulate the problem as a stochastic game. In a stochastic game, at time $t$ each agent $i$ in state $s_t$ takes an action $a_t^i$ according to the policy $\pi^i$. All the agents then transition to the state $s_{t+1}$ and receive a reward $r_t^i$. Stochastic games can be described as as tuple $\langle \mathcal{S}, \mathbf{A}, P, \mathbf{R} \rangle$, where $\mathcal{S}$ is the set of states, and $\mathbf{A} = \{\mathcal{A}^1,\dots,\mathcal{A}^m \}$ is the joint action space consisting of the set of each agent's actions, where $m$ is the number of agents. The reward functions 
$\mathbf{R} = \{\mathcal{R}^1,\dots,\mathcal{R}^m \}$ describe the reward for each agent $\mathcal{S} \times \mathbf{A} \rightarrow \mathbf{R}$. The transition function 
\mbox{$P: \mathcal{S} \times \mathbf{A} \times \mathcal{S} \rightarrow [0,1]$} describes how the state evolves in response to all the agents' collective actions. 
Stochastic games are an extension to Markov Decision Processes (MDPs) that generalize to multiple agents, each of which has its own policy and reward function. 

Let $x_t^i$ be the local state of a single agent.
We will refer to the sequence of the local states, actions, and rewards for a single agent as a trajectory $\mathbf{\tau}^i = \{(x_1^i, a_1^i, r_1^i), \ldots, (x_T^i, a_T^i, r_T^i)\}$ over a horizon $T$. 

The goal is to learn an optimal ego-agent policy $\pi^{*ego}$ where at every point in the learning process $\pi^{ego}\in \bm{\Pi}^{ego}$.   

\section{PROPOSED PIPELINE}
\begin{figure}
\centering
\includegraphics[height=1.3in]
{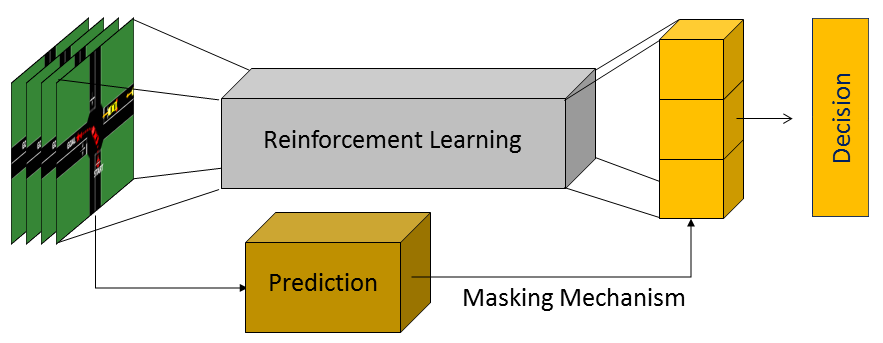}
\caption{Proposed pipeline.}\label{fig:pipeline}
\end{figure}

To achieve safe learning we propose the following pipeline presented in Figure \ref{fig:pipeline}. In this pipeline, perception provides input to both an RL network and a prediction module. The prediction module masks undesired actions at each time step. 

Given predictions, we can mask actions that result in unsafe behaviors. Masking unsafe behaviors in RL has also very recently been proposed in the RL community when states can be mapped to linear temporal logic \cite{alshiekh2018safe}. This pipeline provides us with a mechanism to explore the safe subspace of an agent's possible behaviors during training and also guarantees that RL picks safe decisions during execution. 

\section{PREDICTION}

We propose using prediction models to mask unsafe actions from the agent, and then allow the agent to freely explore the safe state space using traditional RL techniques. Probabilistic predictions serve as an approximation to the computationally expensive task of identifying safe trajectories.  

RL algorithms have been proposed for stochastic games without any restrictions on safety \cite{littman1994markov,hu2003nash}. However, in order to ensure that the agent never takes an unsafe action, we must check not only that a given action will not cause the agent to transition to an unsafe state in the next time step, but also that the action will not force the agent into an unsafe state at some point in the future. 
Note that this is closely related to the credit assignment problem, but the risk must be assigned prior to acting. One might imagine that ensuring the agent safely avoids all dangerous situations requires branching through all possible action combinations for a fixed time horizon $T$. Brute force implementations would result in an intractable runtime of $O\left( |\mathbf{A}|^{T} \right)$, 
where $|\mathbf{A}|= |\mathcal{A}^1|\times \dots \times |\mathcal{A}^m|$. Indeed it has been shown that even for the more restricted case of MDPs, identifying the set of safe policies is NP-Hard \cite{moldovan2012safe}. For this reason we look at efficient approximations for restricting our exploration space.

To reduce the complexity 
we assume the actions at each time step are components of a high-level action (known as options in the RL literature, and intentions in the autonomous driving literature). This has the effect of collapsing the branching factor of time associated with the exponential complexity. The cost of this approximation is that, for the fixed horizon, each agent is restricted in their ability to react and interact with other agents. 

To accommodate the breadth of low-level action sequences that correspond to a single high-level action and also to allow for a bounded level of interaction, we make each high-level action a probability distribution over functions $f$. First we describe the trajectory of agent $i$ in terms of high-level actions 
$ p(\mathbf{\tau}^i) \approx \prod_{j=1}^{|h^i|} p(h^{i}=j) p_{h^{i,j}}(x_1^i, \dots, x_T^i)$. Here $j$ indexes the high-level action $h$. Then we describe the functional local-state update as $x_{t+1}=f^{i,j}(x_t)+ \epsilon_t$ where we model the noise as a Gaussian distribution $\epsilon_t = \mathcal{N}(0,\sigma_t)$.  
This means that the updated local state has a corresponding mean and variance. 

Within the fixed time horizon, each agent takes a single high-level action. The variance acts as a bound that encompasses the variety of low-level actions that produce similar high-level actions. Additionally, we will use the variance to create safety bounds. These bounds allows for a bounded ability of each agent to react to other agents without violating our safety constraints. This can be thought of as selecting an open-loop high-level decision followed by subsequent bounded closed-loop low-level corrections.
Note that by restricting an agent's ability to interact and limiting each agent to a restricted set of high-level actions, we are ignoring the existence of many pathological cases that may arise in an adversarial setting. 

Given the assumption of high-level actions that follow a distribution, satisfying safety constraints can be computed in $O(|\mathbf{H}|T)$ where $|\mathcal{H}^i|$ is the number of high level actions available to agent $i$ and $|\mathbf{H}|= |\mathcal{H}^1|\times \dots \times |\mathcal{H}^m|$. This is still expensive for problems with a large number of actions or agents.

A further simplifying assumption arises when we assume an agent's action space is unimodal. This is the case when we assume the agent has a single action (e.g. a constant velocity assumption) or we make a hard prediction of the most probable action. This reduce the time complexity of a forward safety-checking prediction to $O(mT)$.

\section{SAFETY GUARANTEES}
One might suspect that our relaxations make it difficult to provide any safety guarantees. It does greatly limit the strength of the guarantees we can make, however we can still provide probabilistic guarantees on safety.

From Chebyshev's inequality we can state that the likelihood of an agent $i$ taking action $j$ leaving its safety margins $k\sigma^{i,j}$ is $p[|\mathbf{\tau}^{i,j}-\mathbb{E}(\mathbf{\tau}^{i,j})| \ge k \sigma^{i,j}] \le \frac{1}{k^2} $. Where we define $|\mathbf{\tau}^{i,j}-\mathbb{E}(\mathbf{\tau}^{i,j})| \equiv \max_k|x^{i,j}_k - \mathbb{E}(x^{i,j}_k)|$. Note that we use the weaker Chebyshev inequality for our bounds since according to the Fisher$-$Tippett$-$Gnedenko theorem the $\max$ operation results in a distribution that is not Gaussian. 

Since we generally only care about one-sided error (e.g. if the traffic car is further away than predicted, we do not risk a collision) we can shrink the error by a factor of two. 
$p[\mathbf{\tau}^{i,j}-\mathbb{E}(\mathbf{\tau}^{i,j}) \ge k \sigma^{i,j}] \le \frac{1}{2k^2} $. 

In our model, we assume our safety margins create an envelope for an agent's expected trajectory. Collecting sufficient samples of independent trials, we can assume the predicted trajectory roughly models the reachable space of the agent. Now we probe the independence of trials. In expectation the agent follows the mean, but on each trial the deviations are likely not a purely random process, but are biased by a response to other agents. 

In the autonomous driving literature it is assumed that agents behave with self-preservation \cite{lefevre2014survey}. We will assume that the measured distribution of the trajectory is the sum of two normally distributed random processes: the first associated with the agent's control and the second a random noise variable.   The measured variance of a trajectory $\sigma_M^2$ is the sum of controlled $\sigma_c^2$ and noise $\sigma_n^2$ variances. We can express these relative to the measured standard deviation of the trajectory as $\alpha_c \sigma_M$ and $\alpha_n \sigma_M$ where $\alpha_c^2+\alpha_n^2 = 1$ 
If we assume an agent controls away from the mean by $\kappa_c \alpha_c \sigma_M < k\sigma_M$  the probability that an agent leaves its safety margin is $p[\mathbf{\tau}^{i,j}-\mathbb{E}(\mathbf{\tau}^{i,j}) \ge \kappa_n \alpha_n \sigma_M] \le \frac{1}{2\kappa_n^2} $, where $\kappa_n = \frac{k+\kappa_c \alpha_c}{\alpha_n}$. 
To put this in concrete terms for an autonomous driving scenario, if we assume a 5m measured standard deviation, 4m control standard deviation, 3m noise standard deviation, safety margin of $3\sigma_M$, and control action of $2\sigma_c$, the resulting safety margin is $7.6\sigma_n$. This analysis neglects any corrective controls of the ego agent. Applying the union bound and assuming a fixed $\kappa_n$ for notational clarity, we can achieve our desired confidence $\delta$ by satisfying 
\begin{eqnarray}
\frac{m}{2\kappa_n^2}<\delta \enspace.
\end{eqnarray}


\section{APPLICATION TO AUTONOMOUS DRIVING}

There are many works in the autonomous driving literature that look at risk-averse driving \cite{damerow2015risk} and risk assessment \cite{lefevre2015intention,damerow2015risk,brechtel2011probabilistic}. 
Prediction is often used for safety in autonomous driving and 
accurate prediction models are a current topic of research in the autonomous driving community \cite{lefevre2014survey}. Simpler models are built upon kinematic motion models \cite{schubert2008comparison} with added uncertainty estimates to allow for errors in the measurements and assumptions \cite{carvalho2014stochastic}. These methods are limited in that the Gaussian probability models and kinematic transitions assume cars roughly follow a known trajectory. More sophisticated models allow for multiple maneuvers \cite{aoude2011mobile} which can be done by including road information (either heuristically or learned for particular intersections \cite{streubel2014prediction}) to allow for multiple possible maneuvers. More recent work in vehicle prediction is starting to consider the interactions between multiple vehicles \cite{agamennoni2012estimation,kuefler2017imitating}.

Related work has looked at learning policies for intersection handling \cite{song2016intention,gindele2013learning,isele2018aaai,bouton2017belief} however these approaches are restricted to simulation and do not investigate the issue of preserving safety throughout the learning process under uncertainty. 
Using prediction as a safety constraint does not necessarily require additional learning. 
Accurate prediction models could be sufficient to create behaviors that enforce safety constraints. 
A robust vehicle prediction module could itself be used to safely navigate intersections: 
\begin{enumerate}
	\item Predict the movement of the ego car entering the intersection in conjunction with forward predictions of all other vehicles.  
	\item If a collision is predicted, wait. 
	\item Otherwise, go. 
\end{enumerate}
This however assumes a fixed behavior for the ego car - a single acceleration profile, a set time allowance to enter the intersection, and a set safety buffer to leave between cars.  

Limiting the behavior of the autonomous vehicle to a one-size-fits-all motion is likely to lead to sub optimal behavior. Certain intersections may call for more aggressive accelerations to prevent excessive waiting. And the ability of other agents to interpret our actions can be just as important to safety as leaving sizable margins to allow for uncertainty. 
This work sets up a methodology by which we can explore the more nuanced aspects of decision making.
As an example we consider learning a model that minimizes traffic disruptions.



\begin{figure*}[thpb!]
\centering
\includegraphics[height=1.6in]{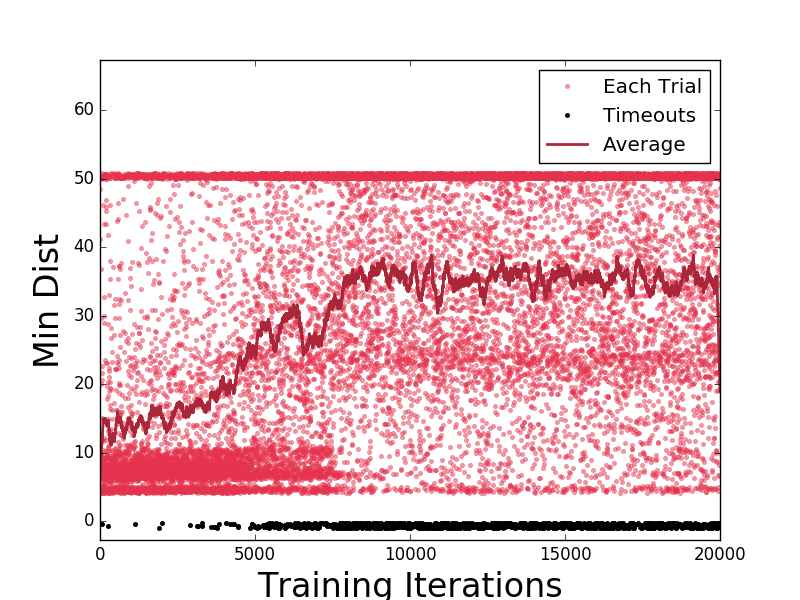}
\includegraphics[height=1.6in]{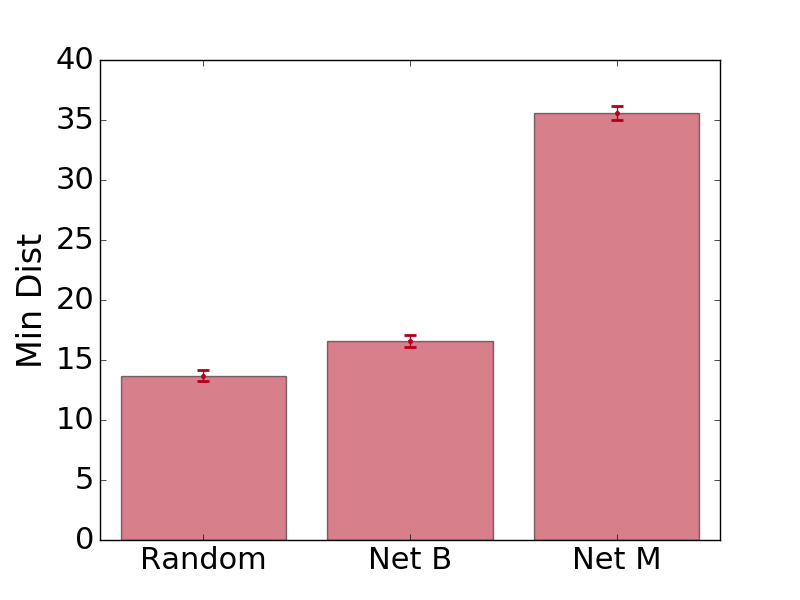}
\includegraphics[height=1.6in]{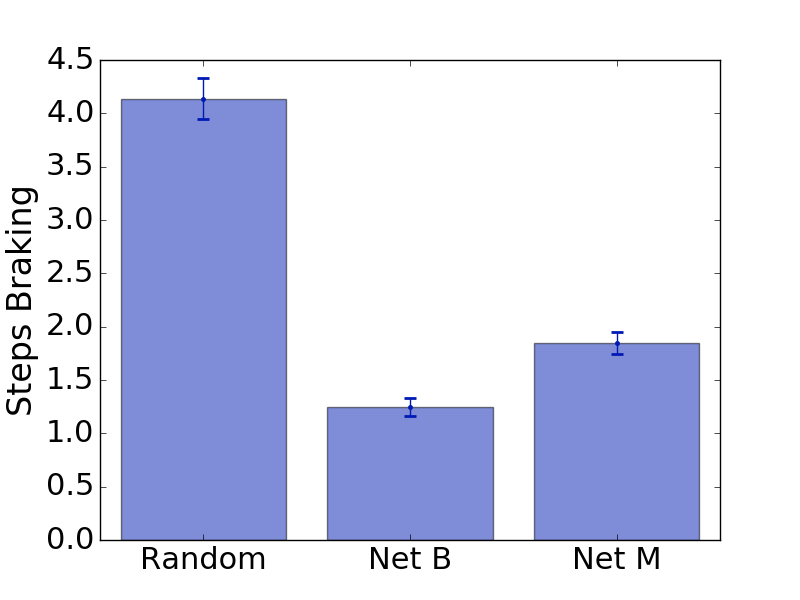}
\caption{\textbf{Left:} Minimum distance to traffic vehicles throughout the training process. \textbf{Center:} Comparison of the minimum distance to traffic vehicles using different policies. Net B is trained to minimize braking, net M is trained to maximize the minimum distance. \textbf{Right:} The amount of time steps traffic cars spend braking. We assume the ego vehicle is responsible for all traffic braking. }\label{fig:comb}
\end{figure*}

\section{EXPERIMENTS}\label{sec:experiments}
To demonstrate how prediction can be used as a safety constraint, we use deep Q-learning networks (DQNs) to learn policies that optimize aspects of intersection handling on autonomous vehicles. We consider two objectives. The first objective is to learn an adaptive stand off which seeks to increase the safety margin without compromising the ability to make the turn given a fixed time window. The second model looks at minimizing the disruption to other vehicles while navigating the intersection in the given time.

\subsection{Prediction}
We model traffic vehicles using a constant velocity assumption based our Kalman filter estimates of the detected vehicle. Each vehicle is modeled with a fixed 2m uncertainty in detection. An additional uncertainty per time step is accumulated forward in time following a quadratic curve which was fit to data collected from errors in the forward velocity assumption targeting a margin of six standard deviations. This allows the model to make allowances for some accelerations and braking of the traffic vehicles. 
The ego car has similar forward predictions of its behavior based on the target trajectory and three potential acceleration profiles. The prediction errors are smaller for the ego car, since the intentions are known in advance. 
At each time step, going forward in time until the ego car has completed the intersection maneuver, the predicted position of the ego car is compared against the predicted position of all traffic cars. If an overlap of the regions is detected, the action is marked as \emph{unsafe}. Actions that are marked as \emph{safe} are passed on to the network as permissible actions. If there are no permissible actions, or the network chooses to wait, the system waits at the intersection. Otherwise the vehicle moves forward with the selected acceleration until it reaches its target speed.  

\subsection{Simulation}
Experiments were run using the Sumo simulator \cite{sumo}, which is an open source traffic simulation package. 
To simulate traffic in Sumo, users have control over the types of vehicles, road paths, vehicle density, and departure times. Traffic cars follow the Intelligent Driver Model (IDM) \cite{idm-def} to control their motion. In Sumo, randomness is simulated through driver imperfection models (based on the Krauss stochastic driving model \cite{krauss1998sumo}). The simulator runs based on a predefined time interval which controls the length of every step. For our experiments we use 0.2 second time step. 

Each lane has a 30 mile per hour (13.4 m/s) max speed. The car begins from a stopped position. 
The maximum number of steps per trial is capped at 100 steps,  which is equivalent to 20 seconds, starting from the first time prediction says a safe action is possible. This guarantees that a safe action is always possible in the allotted time. 
We use a 0.1 probability that a vehicle will be emitted per second to set the traffic density for our experiments.

The DQN architecture is modeled after the network presented in \cite{isele2018icra}. The simulator is designed to see cars 100m in either direction. This was selected to correspond to 25mph intersections. If we assume that it takes roughly 5 seconds to enter an intersection from a stopped position (measured from human demonstrations), we would like to detect traffic at a minimum of 55m from the intersection. This minimum increases to 75m when we allow for traffic vehicles that travel slightly above the speed limit. The remaining 25m provide an added buffer to allow for accurate detection and tracking. The IBEO sensors we use on the real vehicle are specified at 200m max range. The representation bins the traffic car positions into 26 bins per lane. Each lane is depicted as a separate row. Each spatial pixel, if occupied, contains the normalized real valued heading angles, velocity, and binary indicator. 

The network has four outputs corresponding to \emph{wait} and \emph{go} commands where \emph{go} can select from three different accelerations (0.5, 1.0, and 1.5 $m/s^2$). 
The network is optimized using the RMSProp algorithm \cite{tieleman2012lecture}. 


Each network was trained on $20,000$ simulations.
When learning a network that seeks to minimize braking. The per trial reward is $+1$ for successfully navigating the intersection with a $-0.1$ penalty applied for every time step a traffic vehicle was braking. 

When learning a behavior that seeks to maximize the safety margin, the per trial reward is 

\[
    r = \Bigg\{\begin{array}{lr}
        0.1(d-10), & \text{if success } \\
        z, & \text{if timeout }
        \end{array}
\]

Where $d$ is the minimum distance the ego car gets to a traffic vehicle during the trial. $d$ can be a maximum of 50m and the minimum observed distance during training is 4m. We conduct experiments with different $z$ values $z=\{-1,-5,-10\}$ to study the affect on timeouts.  

To evaluate the learned models we use two metrics: the average number of time steps per trial a traffic vehicle is braking, and the minimum distance between the ego car and the closest traffic car per trial. Statistics are collected over 1000 trials. Since both objectives could be improved by increasing the safety margin, we also conduct an experiment where we increase the safety margin of a rule-based only agent to ensure the learned policy gives us an improvement. 


\subsection{Real vehicle}
We train in simulation and verify the learned policy on an autonomous vehicle. We collected data from an autonomous vehicle in Mountain View, California, at an unsigned T-junction, where the objective is to make a left turn. A point cloud, obtained from six IBEO Lidar sensors, is first pre-processed to remove points that reside outside the road boundaries. A clustering of the Lidar points with hand-tuned geometric thresholds is combined with the output from three Delphi radars to create the estimates for vehicle detection. Each vehicle is tracked by a separate particle filter. Figure \ref{fig:car} depicts an intersection where our algorithm was evaluated. We use the network trained to maximize the safety margin.

\section{RESULTS}

\begin{figure}
\centering
\hspace{-20pt}
\includegraphics[clip,trim={30pt 30pt 30pt 30pt},height=2.0in]{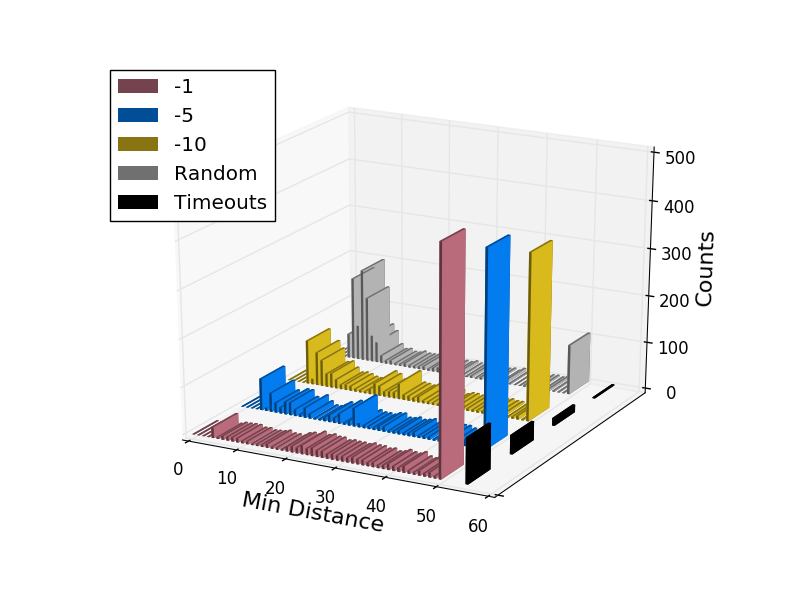}
\caption{Comparison of the effect of the timeout penalty on the network trained to maximize the minimum distance. Timeouts are shown in black.}\label{fig:hist}
\end{figure}

Using prediction as a constraint, we trained a DQN to minimize distance to other traffic vehicles. There were no recorded collisions during the entire training process. The left plot in Figure \ref{fig:comb} shows how the minimum distance to traffic vehicles changes throughout the learning process when $z=-1$. Timeouts are shown as having a distance of -1 and are colored black. The minimum distance of each individual trial is plotted as a point. The moving average using a sliding window of 200 trials is shown in dark red. We see the concentration of points with a distance of less than 10m disappears at around 7000 training iterations, and there is an increased density in the region of 20m. Training does increase the number of timeouts. The extent of this can be changed by adjusting the penalty for timeouts, see Figure \ref{fig:hist}. Note that larger penalties produce larger gradients which can have an adverse affect on the learning process so there is a limit to how large the penalty can be set.   

\begin{figure}
\centering
\hspace{-20pt}
\includegraphics[height=1.75in]{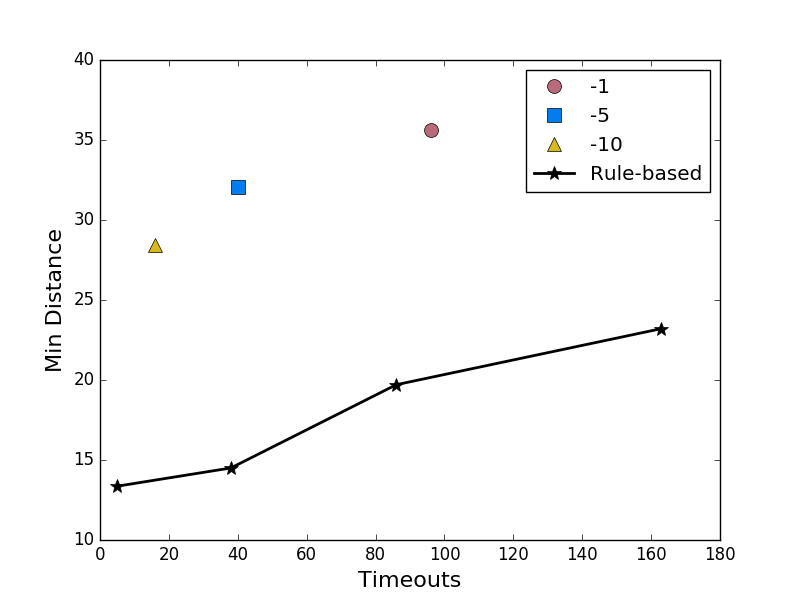}
\caption{Comparison between the networks trained with different timeout penalties, and a rule-based method with a fixed safety margin. The safety margin for the rule-based method is varied across trials.}\label{fig:compare}
\end{figure}

The number of trials that have a minimum distance of 50m or more also increases. This is more clear in Figure \ref{fig:hist} where we show a histogram of the distances before and after training. Figure \ref{fig:compare} shows that naively increasing the safety margin with a rule-based strategy using prediction gives much worse performance compared to the learned networks. 


The network we trained to minimize braking should leave a large margin when moving in front of a car, however it may come up very close behind a car. We see that this is the case in center plot of Figure \ref{fig:comb}. As expected the network trained to maximize the minimum distance (Net M) greatly increases the average minimum distance. The network trained to minimize braking does achieve a larger average distance than a random policy acting in the safety constrained prediction framework, but the difference is much less pronounced. 
The network that was trained to maximize distance should also reduce braking. This result can be seen in right plot of Figure \ref{fig:comb}. Again we see the network specifically designed to minimize braking is better at satisfying the objective. 
The fact that these two objectives produce different behaviors makes sense. If we think in terms of the gap between two cars, it would be optimal to be in the middle of the gap if maximizing distance and the front of the gap if minimizing braking of other vehicles.

Qualitatively, we observe that the network trained to maximize the safety margin will often add short delays after prediction determines the situation is safe if traffic is sparse. In cases where traffic is more dense, the network is more likely to move the moment an opportunity presents itself.  

\section{CONCLUSION}

In this work we present a framework for safe RL using predictions to mask unsafe actions. We apply this methodology to an autonomous driving domain to learn policies that improve the performance of unsigned intersection handling. Specifically we look at 1) minimizing disruption to other vehicles and 2) maximizing safety margins while still navigating the intersection in a fixed time window.  

While the safety guarantees we can make using prediction are not as strong as other approaches proposed in the literature, the framework is more general and likely more applicable to many real world applications. Since we are masking actions, some of which we know to be safe in order to provide safety margins when dealing with uncertainty, the final policies are possibly suboptimal. This suggests open problems both related to developing more sophisticated prediction modules and a more careful characterization of the regret associated with them.  

\bibliographystyle{IEEEtran}
\small{
\bibliography{refs}
}

\end{document}